%% file: cameraready.tex
\pdfoutput=1

\documentclass[11pt]{article}

\usepackage[final]{acl}
\usepackage{comment}
\usepackage{times}
\usepackage{latexsym}
\usepackage{booktabs}
\usepackage{multirow}
\usepackage{multicol}
\usepackage[table]{xcolor}
\usepackage{adjustbox}
\usepackage{hyperref}
\usepackage{amsmath}
\usepackage[T1]{fontenc}
\usepackage{arydshln}
\usepackage{enumitem}
\usepackage{subcaption}   
\usepackage{graphicx}
\usepackage{enumitem}
\usepackage{xcolor}  

\usepackage[utf8]{inputenc}

\usepackage{microtype}

\usepackage{inconsolata}

\title{Scaling Low-Resource MT via Synthetic Data Generation with LLMs}

\author{
  \textbf{Ona de Gibert\textsuperscript{\raisebox{-0.1ex}{\includegraphics[width=2.5ex]{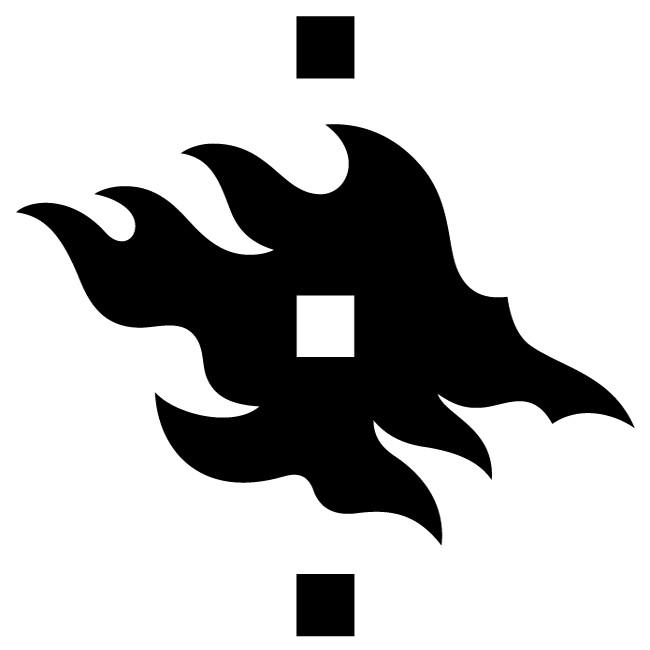}}}} \quad
  \textbf{Joseph Attieh\textsuperscript{\raisebox{-0.1ex}{\includegraphics[width=2.5ex]{tabs_and_figs/helsinki.png}}}} \quad
  \textbf{Teemu Vahtola\textsuperscript{\raisebox{-0.1ex}{\includegraphics[width=2.5ex]{tabs_and_figs/helsinki.png}}}} \quad
  \textbf{Mikko Aulamo\textsuperscript{\raisebox{-0.1ex}{\includegraphics[width=2.5ex]{tabs_and_figs/helsinki.png}}}} \quad
  \textbf{Zihao Li\textsuperscript{\raisebox{-0.1ex}{\includegraphics[width=2.5ex]{tabs_and_figs/helsinki.png}}}} \\
  \textbf{Raúl Vázquez\textsuperscript{\raisebox{-0.1ex}{\includegraphics[width=2.5ex]{tabs_and_figs/helsinki.png}}}} \quad
  \textbf{Tiancheng Hu\textsuperscript{\raisebox{-0.1ex}{\includegraphics[width=2ex]{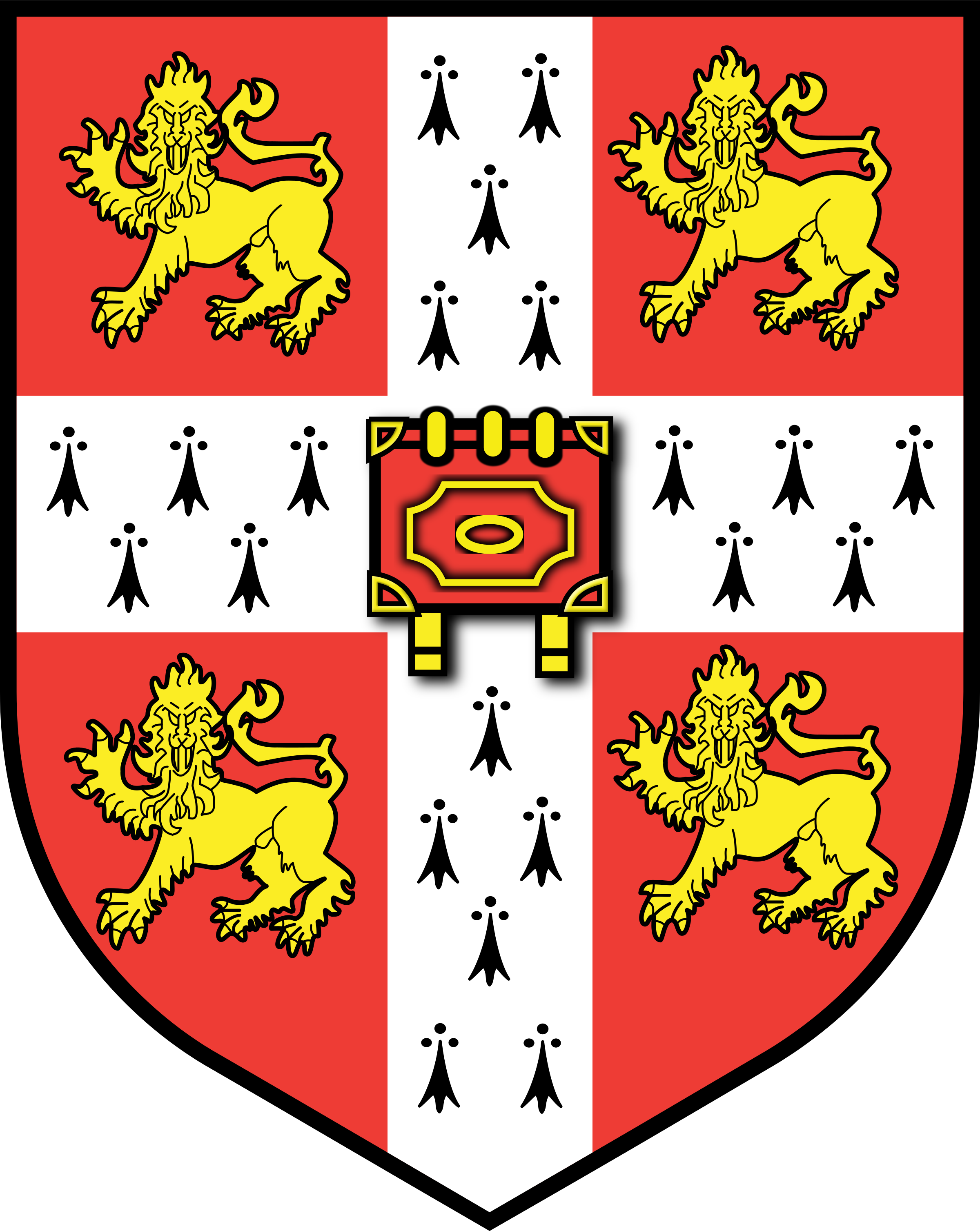}}}} \quad
  \textbf{Jörg Tiedemann\textsuperscript{\raisebox{-0.1ex}{\includegraphics[width=2.5ex]{tabs_and_figs/helsinki.png}}}} \\
    \textsuperscript{\raisebox{-0.1ex}{\includegraphics[width=3ex]{tabs_and_figs/helsinki.png}}} University of Helsinki    \quad \textsuperscript{\raisebox{-0.1ex}{\includegraphics[width=2.5ex]{tabs_and_figs/cambridge.png}}} University of Cambridge \\
   \{first.last\}@helsinki.fi,  th656@cam.ac.uk \\
}

\begin{document}
\maketitle
\begin{abstract}
We investigate the potential of LLM-generated synthetic data for improving low-resource Machine Translation (MT). 
Focusing on seven diverse target languages, we construct a document-level synthetic corpus from English Europarl, and extend it via pivoting to 147 additional language pairs.
Automatic and human evaluation confirm its overall high quality.
We study its practical application by (i) identifying effective training regimes, (ii) comparing our data with the HPLT dataset, (iii) studying the effect of varying training data size, and (iiii) testing its utility beyond English-centric MT. Finally, we introduce {SynOPUS}, a public repository for synthetic parallel datasets. 
Our findings show that LLM-generated synthetic data, even when noisy, can substantially improve MT performance for low-resource languages. 
\end{abstract}
\section{Introduction}

Machine translation (MT) has achieved remarkable success for high-resource languages, but its application to the vast majority of the world's languages remains severely hampered by the scarcity of high-quality parallel corpora.
Traditional data augmentation techniques like back-translation \cite{sennrich2016improving} and pivoting \cite{costa2018english,cheng2019joint} preserve the human-written target and synthesize the other.
The advent of Large Language Models (LLMs) presents a transformative opportunity, as reflected by the growing number of survey papers on the subject \cite{zhou2024survey,ding2024data,wang2024survey,nadas2025synthetic}. LLM-based synthetic data generation, 
akin to sequence-level knowledge distillation \cite{kim2016sequence}, opens up the possibility of creating vast amounts of training data even where human-translated resources are virtually non-existent.



This raises the question: \textit{Can an MT system trained on LLM-generated data benefit truly low-resource language pairs?} To date, there is little to no systematic investigation of (a) generating large-scale synthetic data using LLMs for low-resource languages, (b) evaluating its intrinsic quality, (c) quantifying its downstream impact when training or fine-tuning modern MT systems.
This paper provides such a systematic investigation. We make the following contributions:
\begin{itemize}[
  leftmargin=1em,  
  itemsep=0.5ex,     
  parsep=0pt,        
  topsep=0.5ex,      
  label=\textbullet  
]
    \item We use GPT-4o\footnote{We use the \texttt{gpt-4o-2024-08-06} model.} to generate a document-level synthetic parallel corpus by forward-translating English Europarl \cite{koehn2005Europarl} into seven diverse low-resource languages.
    \item We assess the corpus quality using both automatic metrics and human evaluation, finding the data to be generally of high quality.
    \item We comprehensively evaluate the utility of this synthetic data by demonstrating that:
    \begin{enumerate}[
      leftmargin=1.5em,     
      itemsep=0.3ex,      
      label=(\arabic*),    
      font=\normalfont    
    ]
        \item Compact MT models trained from scratch solely on this data achieve strong baseline performance 
        (e.g., 49.49 ChrF for English-Georgian, compared to NLLB's 48.31).
        \item Fine-tuning pretrained state-of-the-art (SOTA) systems (OPUS-MT, NLLB-200-1.3B, Llama-3B) consistently yields substantial improvements (e.g., average gains of +2.95 ChrF for NLLB and +20.63 ChrF for Llama-3B).
        \item Our synthetic data is complementary to existing corpora like HPLT, e.g., leading to further ChrF increases of up to +2.79 when combined (for English-Icelandic).
        \item Fine-tuned models that are 
        10-20 times smaller than SOTA models perform similar or better than their large counterparts.
    \end{enumerate}
    \item \textcolor{black}{We study the effect of training data size to investigate the scalability of our approach.}
    \item We extend our dataset into a multi-way parallel corpus via pivoting. As a case study, we demonstrate that Finnish-Somali translation improves by +14.78 ChrF and +21.64 ChrF when fine-tuning OPUS-MT.
    \item To promote reproducibility and future research, we introduce {SynOPUS}, a public repository of synthetic parallel corpora. We also publicly release our dataset, \textcolor{black}{which we publish under a new version of Europarl (v8syn)},\footnote{ \url{https://opus.nlpl.eu/synthetic/Europarl.php} The data is subject to the terms and conditions defined by the \href{https://openai.com/policies/usage-policies/}{usage policies of OpenAI.}} code,\footnote{\url{https://github.com/Helsinki-NLP/low-res-lmt}} and baseline models.\footnote{\href{https://huggingface.co/collections/Helsinki-NLP/scaling-low-res-mt-via-synthetic-data-generation-with-llms-68c91809042d2f34e9d44841}{Helsinki-NLP/scaling-low-res-mt-via-synthetic-data-generation-with-llms}}
\end{itemize}

Our results show that LLM-generated synthetic data, even when noisy, can train competitive MT models from scratch and consistently improves pretrained systems, especially for the least resourced languages in the resource spectrum. Our work demonstrates a clear path towards 
open high-quality MT for underrepresented languages, by harnessing widely available high-resource monolingual corpora and powerful LLMs.

\section{Related Work}

\paragraph{Low-resource MT.}
Low-resource MT targets language pairs with little to no parallel data available  \cite{haddow2022survey}. To mitigate the data scarcity problems, two main lines of research have emerged: (a) transfer learning \cite{zoph2016transfer} and multilingual training \cite{johnson2017google}, and (b) data augmentation \cite{xia2019generalized}. 
First, transfer learning involves using a model trained on a high-resource language as a starting point for training the low-resource language, while multilingual training proposes to train jointly on multiple language pairs  to compensate for the lack of text in a specific language. 
Second, data augmentation proposes to generate synthetic samples to train on, by perturbing, translating or otherwise modifying existing sentences \cite{fadaee-etal-2017}.  Below, we focus on data augmentation and recent work that uses LLMs to generate such data. 

\paragraph{Classical data augmentation.}
The most popular approach for low-resource languages is back-translation, which involves translating the monolingual target-language data into the source language \cite{ko-etal-2021-adapt,khenglawt2024addressing}. 
The reverse process, forward translation of source-side monolingual sentences, has also been explored, and while less common in MT,  proved valuable for LLM pretraining.  For example, \citet{wang2025multilingual} used NLLB to forward‐translate monolingual corpora in nine languages and demonstrated its value for LLM pretraining. This process also relates closely to sequence-level knowledge distillation \cite{kim2016sequence}, where compressing a large model involves training a small \textit{student} model on synthetic data constructed by forward-translating it with the \textit{teacher} model \cite{gordon2019explaining}. 

\paragraph{LLM-based data augmentation.} 
LLMs have opened new avenues for synthetic data generation, driven by their strong performance in low-resource language settings. Several studies assess the translation performance of LLMs: Claude on Yoruba-English \cite{enis2024llm}, Claude on 13 low-resource languages of Mali \cite{dembele2025serendipity}, and GPT-4 on 3 languages \cite{jiao2023chatgpt}.  
These efforts encouraged researchers to use LLMs for synthetic data generation. For instance, 
\citet{oh2023data} explore different prompting strategies to generate synthetic data for German-Korean translation with ChatGPT. 
Our work is most similar to \citet{yang2023neural}, where they exploit data generation for MT between German and Galician with ChatGPT. However, the authors generate source synthetic sentences that are later translated, while we use original English sentences as source data, and experiment on more languages. 

\paragraph{Gap addressed in this work.} Despite the above advances, there is still no systematic study that produces a fully synthetic multi-way parallel corpus with SOTA LLMs for low-resource languages and evaluates that corpus both intrinsically and on downstream MT.  We close this gap by extending Europarl, a multilingual resource with alignments across all the official EU languages, into seven low-resource languages and evaluating its quality and usefulness.

\section{Dataset Construction}

\begin{table*}[ht]
    \centering
    \small
    \begin{tabular}{lccccccc}
        \toprule
        & eu & gd & ka & is & mk & so & uk \\
        \midrule
        n. after segmentation   & 2\,167\,164 & 2\,192\,082 & 2\,504\,071 & 2\,370\,036 & 2\,054\,167 & 2\,373\,145 & 2\,359\,720 \\
        n. after lang. id.      & 2\,160\,061 & 2\,182\,553 & 2\,481\,357 & 2\,362\,411 & 2\,044\,219 & 2\,364\,985 & 2\,351\,562 \\
        n. aligned sentences    & 2\,138\,713 & 2\,164\,999 & 2\,317\,070 & 2\,348\,030 & 2\,027\,406 & 2\,353\,915 & 2\,341\,706 \\
        \bottomrule
    \end{tabular}
    \caption{Statistics of the different post-processing steps described in Section \ref{sec:data_pro} for each language. 
    }
    \label{tab:language_stats}
\end{table*}

Our goal is to study real-world cases instead of selecting common language pairs in an artificially constructed low-resource scenario. We conduct a preliminary experiment to help us select
the languages to prioritize (Section~\ref{sec:language-selection}). We then forward-translate the English Europarl corpus (Section~\ref{sec:data-generation}), and, in a final post-processing step, we 
filter out noise to ensure high-quality translations (Section~\ref{sec:data_pro}). Finally, we expand our dataset via pivoting to all languages of Europarl (Section \ref{sec:multieuroparl}).

\subsection{Language 
Selection}
\label{sec:language-selection}

We start by selecting a small set of low-resource languages for which GPT-4o can produce usable translations. To do so, we begin with a list of 204 European minority languages\footnote{The list is derived from \url{https://en.wikipedia.org/wiki/Regional_and_minority_languages_in_Europe}.} and retain only the 39 languages that are supported by the FLORES+ benchmark \cite{goyal2022flores}. 
%
For each of the 39 languages, we prompt GPT-4o to produce translations of (i) 100 random samples from the FLORES+ development set and (ii) 20 five-sentence chunks to simulate paragraph-level translation. 
We specify the script of the target language in the prompt.\footnote{Our initial experiments suggested that GPT-4o occasionally produces translations using a script different from that used in the FLORES+ dataset. This issue was the most prominent in Serbian, which uses both Cyrillic and Latin scripts.}

To contextualize GPT-4o's performance against existing well-performing translation models, we translate the same datasets with EMMA-500 \citep{ji2024emma}, using both zero-shot and 3-shot settings, by selecting 3 unused examples from FLORES+.
Additionally, we 
compare the results to the best available OPUS-MT model \citep{tiedemann2024democratizing} per language, selected from the OPUS-MT Dashboard \cite{tiedemann2023opus}. We compare the performance of the three systems using ChrF \cite{popovic2015chrf}.\footnote{\texttt{nrefs:1|case:mixed|eff:yes|nc:6|nw:0|space:no| version:2.5.1}}
Appendix~\ref{sec:pilot_eval} (Table~\ref{tab:pilot_eval}) 
presents the results of the pilot evaluation.
We proceed to 
select seven languages: Basque (\textit{eu}), Scottish Gaelic (\textit{gd}), Icelandic (\textit{is}), Georgian (\textit{ka}), Macedonian (\textit{mk}), Somali (\textit{so}), and Ukrainian (\textit{uk}). 
\textcolor{black}{We select these languages based on the linguistic diversity, low-resource coverage, model performance, and our practical interest.}

\subsection{Synthetic Data Generation}
\label{sec:data-generation}

We use the English Europarl\footnote{To the best of our knowledge, the Europarl corpus is not subject to any copyright restrictions.} \cite{koehn2005Europarl} as the source for generating the synthetic dataset. Europarl, which is derived from the proceedings of the European Parliament, offers well-defined document boundaries and is multi-way parallel across 21 European languages. We leverage the metadata within the Europarl corpus to segment the data into paragraphs in a way that each generated translation can be matched back to its exact English source, preserving the multi-way parallel structure.
Paragraphs are sent in bulk to the OpenAI's Batch API\footnote{\url{https://platform.openai.com/docs/guides/batch}}. 
%
We instruct the model to generate translations for source-target language pairs using the following prompt:
\begingroup
\addtolength\leftmargini{-0.2in}
\begin{quote}
    \emph{This is an English to \texttt{TARGET} translation, please provide the \texttt{TARGET} translation to this sentence in \texttt{SCRIPT} script. Do not provide any explanation or text apart from the translation.}
\end{quote}
\endgroup

For instance, in the English-Ukrainian direction, we set the target language (\texttt{TARGET}) to \textit{Ukrainian}, and the script information (\texttt{SCRIPT}) to \textit{Cyrl}.
\textcolor{black}{We use script identifiers from the FLORES+ language codes, such as \textit{ukr\_Cyrl}.}


\subsection{Data Post-Processing}
\label{sec:data_pro}
After translating the data with GPT-4o, we align the translated sequences with the original English sentences to produce parallel datasets. 
To produce aligned sentence pairs, we must first segment the paragraphs into individual sentences.
For every language except Georgian, we use a sentence splitter from the Moses package~\citep{koehn-etal-2007}, selecting the language-specific system whenever it exists. Otherwise, we rely on the 
settings for the closest available language. 
For Somali we use the fallback to English, which seems to perform reasonably well.
For Georgian we apply WtP~\citep{minixhofer-etal-2023} with the \texttt{sat-3l-sm} model for sentence segmentation.

Because of the inherent noise in the translation process, and because of their tendencies to produce hallucinations, the LLMs may make errors in translation. To filter such cases, we apply language identification using \texttt{heliport}\footnote{\url{https://pypi.org/project/heliport/}}, which is based on the HeLI-OTS language identification models~\citep{jauhiainen-etal-2022}, 
to every generated segment after sentence segmentation, and discard those that are not classified correctly. On average, this step removes only about 0.45~\% of sentences in each language. 

Lastly, we align the cleaned target sentences with their English counterparts using the Yasa alignment tool~\citep{lamraoui-langlais-2013}, while preserving document boundaries so each sentence can be matched to its document and sentence identifiers.
The resulting corpus contains 2--2.3 million aligned sentences per language pair. 
The data statistics are presented in Table~\ref{tab:language_stats}.

\begin{figure*}[h!]               
  \centering
  \begin{subfigure}[t]{0.32\textwidth}
    \includegraphics[width=\linewidth]{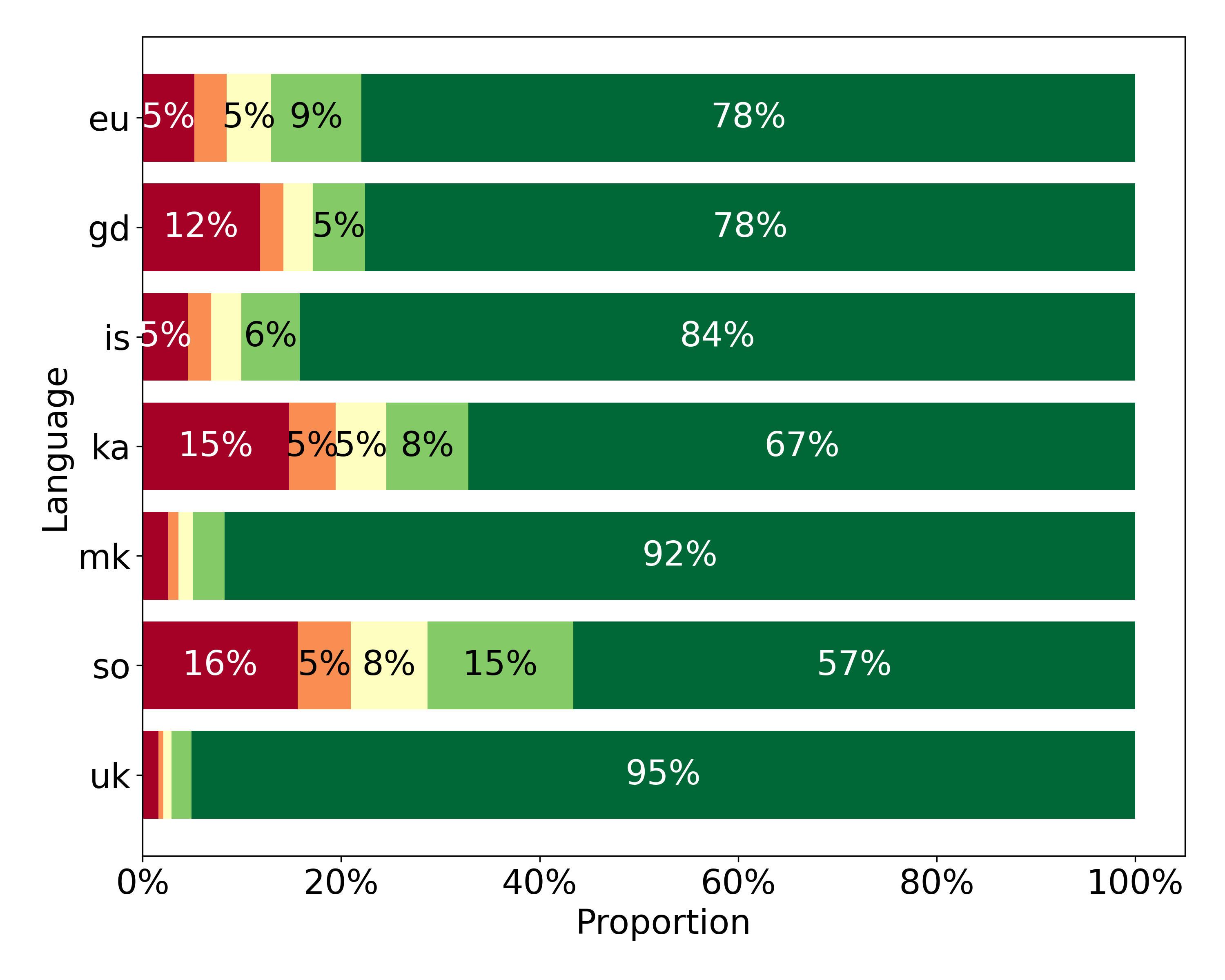}
    \caption{Bicleaner-AI scores.}
    \label{fig:bicleaner}
  \end{subfigure}
  \hfill                    
  \begin{subfigure}[t]{0.32\textwidth}
    \includegraphics[width=\linewidth]{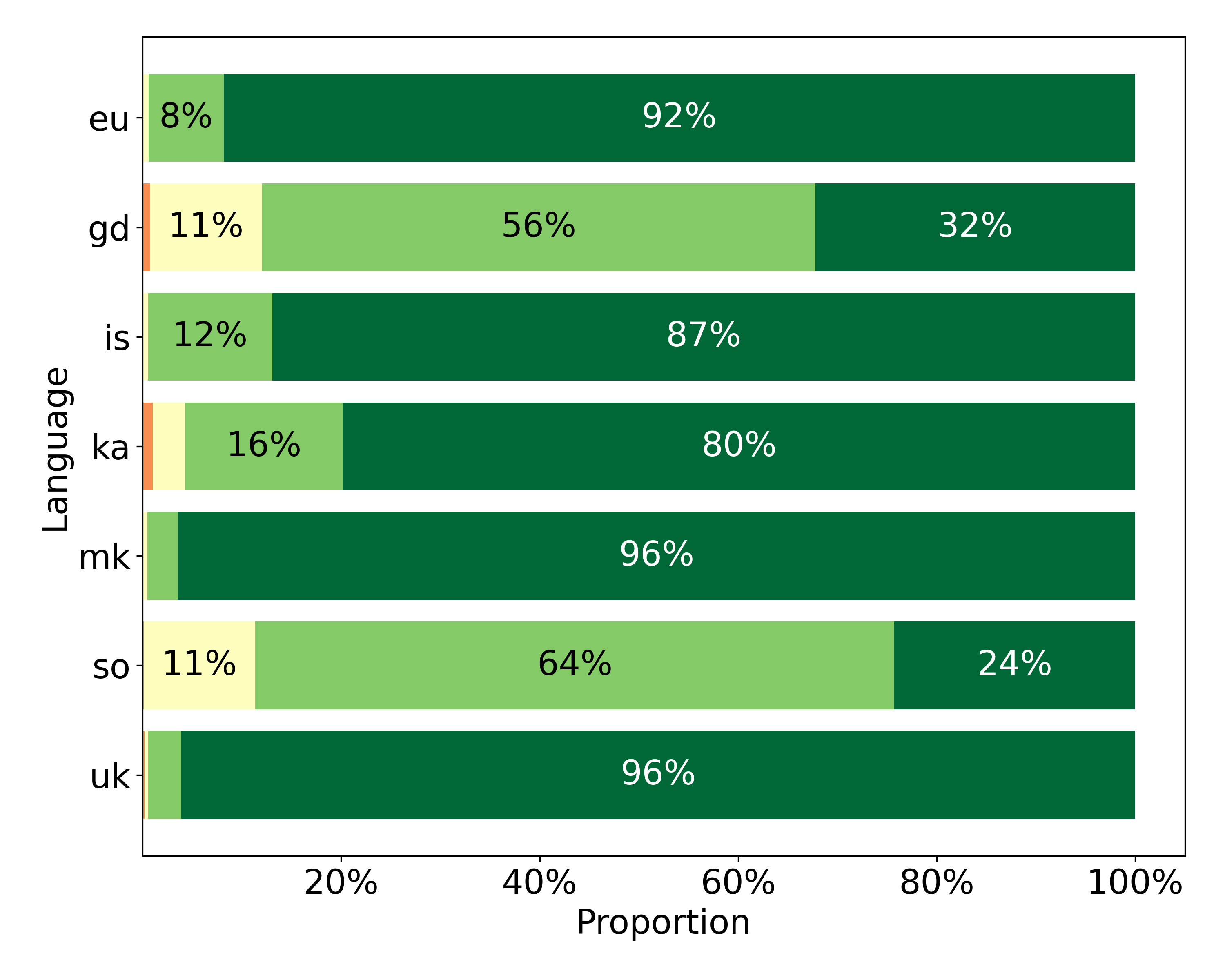}
    \caption{COMET-Kiwi scores.}
    \label{fig:comet}
  \end{subfigure}
  \hfill
  \begin{subfigure}[t]{0.32\textwidth}
    \includegraphics[width=\linewidth]{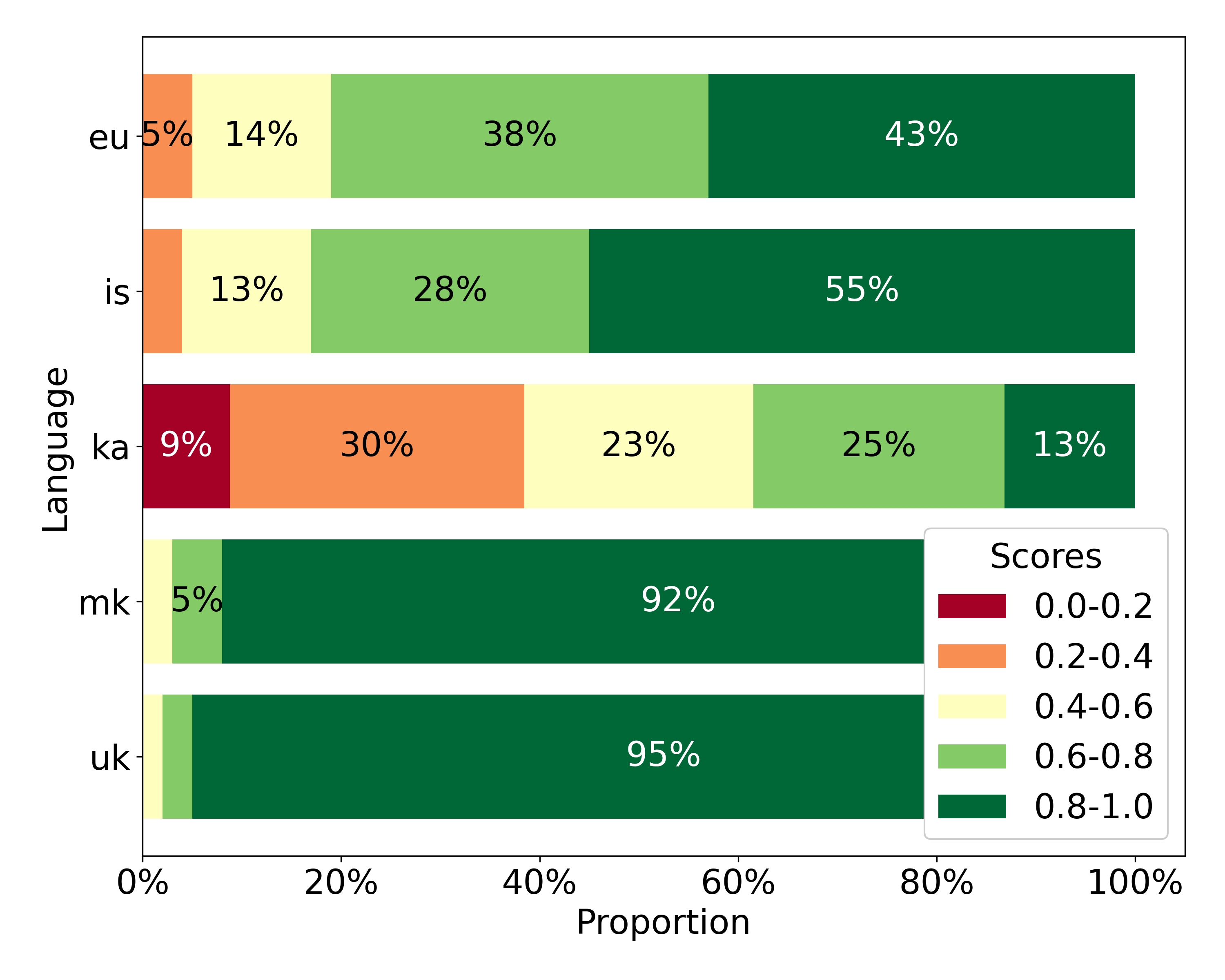}
    \caption{Human evaluation scores.}
    \label{fig:hum_eval}
  \end{subfigure}

  \vspace{-0.5em}         
  \caption{Distribution of scores of the dataset quality analysis.
    Each bar represents the proportion of sentence pairs falling within different score intervals. We normalize human evaluation scores to use the same scale across plots.}
\end{figure*}

\subsection{MultiEuroparl: a Multi-Way Parallel Document-Level Corpus}
\label{sec:multieuroparl}

An important design decision in our experiments was the focus on an inherently multi-way parallel dataset. 
Since all languages are aligned through English, we can use it as a pivot to project synthetic translations onto existing alignments.

In order to achieve that, we preserve sentence, paragraph, and document IDs from the original dataset during translation. 
Then, translated paragraphs are sentence-aligned to their input paragraphs and the alignment of English sentences to other existing languages in the original Europarl corpus are retrieved from OPUS.
A minor complication is that sentence alignment is not one-to-one in all cases. 
We expand alignments by including neighboring sentence pairs until we get a match.
In the worst case, this would cover the entire paragraph but, luckily, the data is quite well-behaved and aligns rather nicely also across language pairs. 


Using the procedures above, we are able to create 147 new language pairs added to the original Europarl corpus, while keeping document information.
\textcolor{black}{Europarl has multi-way parallel corpora originally available in 21 languages. We add 7 new languages to the existing 21, yielding additional $21 \times 7 = 147$ language pairs when pairing each new language with all the existing ones.} 
All of the language pairs are now available as training data for non-English-centric MT, a valuable source that comes for "free" due to the multilinguality and metadata of the source data. We study the usefulness of this data in Section \ref{sec:finnish}.

\section{Dataset Quality Analysis}

In this section, we delve into the quality of the generated low-resource data. We first conduct a quantitative analysis (Section \ref{sec:bicleaner}) producing numerical scores for each sentence pair and then proceed to ask native speakers of the target languages to rate a subsample of the dataset (Section \ref{sec:hum_eval}). Finally, we compute inter-annotator agreement scores and correlation metrics.

\subsection{Quantitative Analysis}
\label{sec:bicleaner}
To evaluate the quality of the generated parallel dataset, we compute two neural metrics at the segment level:  Bicleaner-AI\footnote{We use the 
\texttt{\href{https://huggingface.co/bitextor/bicleaner-ai-full-large-en-xx}{bitextor/bicleaner-ai-full-large -en-xx}} model.} \cite{zaragoza-bernabeu-etal-2022-bicleaner} and COMETKiwi \cite{cometkiwi}. These two metrics are optimized for different tasks and therefore behave differently: Bicleaner-AI is a binary classifier trained to determine whether two sentences are valid translations of each other.
In contrast, COMETKiwi is a reference-free Quality Estimation (QE) metric based on COMET \cite{rei2020comet}, trained to predict human judgment scores (on a 0–100 scale, normalized to 0-1) for machine-translated sentences. 

Figures \ref{fig:bicleaner} and \ref{fig:comet} show the distribution of Bicleaner-AI scores and COMETKiwi per language pair. Looking at the Bicleaner-AI scores, we observe that over 92\% of the sentences in Ukrainian and Macedonian fall in the highest bin and over 12\% of the sentences for Somali, Georgian and Scottish Gaelic fall in the lowest bin. 
Although the general trend of COMETKiwi is similar to Bicleaner-AI, the results are interpreted differently as COMETKiwi reveals the actual quality of the sentences in the dataset generated.  We can see that more than 85\% of the sentences per language are in the top quality bins, noticing that the sentences with lower quality sentences are in Scottish Gaelic, Somali, and Georgian. 
However, COMETKiwi has not been explicitly trained on any of the languages in our dataset, even though its underlying model, XLM-R \cite{conneau2020unsupervised}, includes them. The fine-tuning for QE was conducted using data from the WMT General Shared Tasks (2017–2020). 
As such, these results are zero-shot and should be interpreted with caution.

\subsection{Human Evaluation}
\label{sec:hum_eval}
To further assess the quality of our synthetic dataset, we conduct a human evaluation for five languages.\footnote{We were, unfortunately, unable to find annotators for Scottish Gaelic and Somali.} For each of language pair, we randomly sample 100 sentences and ask native speakers to evaluate them. 
Each language pair was scored by 1--3 annotators. Following the Direct Assessment (DA) protocol \cite{graham2013continuous} from the WMT2017 Shared Task \cite{bojar2017findings}. Annotators were shown the source sentence and its translation, and were asked to assign a score on a 0--100 scale, using the guidelines provided (see Appendix \ref{sec:guidelines}). 
All ratings were collected through a custom web interface built with Gradio \cite{abid2019gradio}. 

We measure Inter-Annotator Agreement (IAA) using Krippendorff’s alpha (interval level) and compute the z-scored version by normalizing each annotator's scores to account for differences in scoring behavior. Table \ref{tab:hum_eval} shows the results of the IAA, which indicates moderate consistency among annotators.

Figure \ref{fig:hum_eval} presents the results of our human evaluation.
Consistent with the findings in the previous section, Macedonian and Ukrainian stand out with high-quality outputs, where human annotators rated over 92\% of the data within the 80–100 score range. Icelandic and Basque exhibited more variability, with approximately 40\% of data rated as good quality (scores between 60 and 100) and around 30\% considered acceptable (scores between 40 and 60). In contrast, Georgian data was considerably lower, with about 40\% judged by annotators as being of unacceptable quality (scores below 40).


\paragraph{Correlation scores} 
We report Spearman’s rank correlation ($\rho_s$) between Bicleaner-AI  and COMETKiwi scores, and human judgments in Table \ref{tab:hum_eval}.
In general, the correlations with BicleanerAI are weak across most language pairs, suggesting limited alignment.
Georgian shows a relatively higher correlation ($\rho_s = 0.39$), likely due to the greater variance in human scores for this language.
We observe that Bicleaner-AI tends to assign lower scores to samples that received very high ratings from human annotators, indicating a potential underestimation of high-quality translations.
In contrast, correlations are consistently higher for COMETKiwi. This is expected, as COMETKiwi is designed to evaluate translation quality (a task more closely aligned with human judgments).

\begin{table}[t!]
    \centering
    \resizebox{\columnwidth}{!}{%
    \begin{tabular}{lrrrr}
    \toprule
    Pair & Ann. Count & z-IAA & $\rho_s\text{(BicleanerAI)}$ & $\rho_s$ (COMETKiwi) \\
    \midrule
    en-eu & 3  &  0.49 & 0.25 & 0.43\\
    en-is & 2 & 0.53 & 0.27 &  0.43 \\
    en-ka & 1 & -  & 0.39 & 0.64 \\
    en-mk & 1 & -  & 0.21 & 0.21\\
    en-uk & 2  & 0.39 & 0.15  &  0.22\\
    \bottomrule
    \end{tabular}
    }
    \caption{Annotator count (Ann. Count), Inter-Annotator Agreement (IAA), as measured by z-score normalized Krippendorff’s Alpha, and Spearman correlation ($\rho_s$) of the human judgements with the Bicleaner-AI and COMETKiwi scores per language pair.}
    \label{tab:hum_eval}
\end{table}

\vspace{5mm}
Overall, we can conclude that the generated dataset is of fairly good quality, with both automatic and human metrics indicating that most sentence pairs are of good to excellent quality, particularly for Ukrainian and Macedonian. Lower performance is observed for Georgian, Somali, and Scottish Gaelic. This is coherent with the GPT-4o pilot evaluation that we conducted on FLORES+, as GPT-4o performs the best in terms of translation accuracy on Ukrainian and Macedonian, and worse for the rest of the languages (see Appendix~\ref{sec:pilot_eval}, Table~\ref{tab:pilot_eval}).

\begin{table*}[ht]
    \centering
            \begin{tabular}{lrrrrrrrr}
    \toprule
      \multirow{2}{*}{Model} & \multicolumn{7}{c}{Language Pair} &  \multirow{2}{*}{\parbox[t]{1.5cm}{$\#$ Params}} \\
   & en-eu & en-gd & en-is & en-ka & en-mk & en-so & en-uk &  \\
    \midrule
    Synthetic    & 53.00 & 51.10 & 49.91 & 49.49 & 57.72 & 45.10 & 51.71 & 60.6M \\ \hline
   OPUS-MT & 54.99 & 41.60 & 51.97 & 42.69 & \textbf{64.45} & 44.20 & \textbf{60.14} & \multirow{2}{*}{191.6M}  \\
    OPUS-MT-ft      & 55.68 & \textbf{52.07} & \textbf{53.80} & 50.16 & 61.99 & \textbf{46.35} & 56.98 &  \\ \hdashline
    $\Delta$  & +0.69* & +10.47* & +1.83* & +7.47* & $-$2.46* & +2.15* & $-$3.16* &  \\ \hline
     NLLB    & 52.05 & 49.94 & 47.98 & 48.31 & 60.13 & 45.90 & 54.44 & \multirow{2}{*}{1.3B}   \\
    NLLB-ft   & \textbf{56.32} & 51.81 & 52.93 & \textbf{52.75} & 62.32 & 46.14 & 57.13 &   \\ \hdashline
    $\Delta$    & +4.27* & +1.87* & +4.95* & +4.44* & +2.19* & +0.24 & +2.69* &  \\ \hline   
    Llama   & 29.25 & 26.56 & 22.66 & 13.17 & 26.58 & 22.76 & 30.24 & \multirow{2}{*}{3B} \\
    Llama-ft        & 49.85 & 47.01 & 46.12 & 25.06 & 55.60 & 42.31 & 49.68 &    \\ \hdashline
    $\Delta$   & +20.59* & +20.45* & +23.46* & +11.89* & +29.02* & +19.55* & +19.44* &  \\ 
    \midrule
    \textcolor{black}{GPT-4o} & 57.10 & 53.24 & 55.94 &  51.84 & 64.45 & 46.82 & 60.88 &  \\ 
    \bottomrule
    \end{tabular}
\caption{ChrF scores (\%) on seven translation tasks. For each architecture (Synthetic, OPUS-MT, NLLB, Llama), we report the raw ChrF of the base and fine-tuned (ft) models when available, along with the absolute improvement ($\Delta$). The rightmost column shows model size. ``\(^*\)'' indicates a significant difference ($p<0.05$) between base and fine-tuned models, based on paired \textit{t}-test and bootstrap resampling (5{,}000 iterations).}

    \label{tab:chrf_results}
\end{table*}

\section{Leveraging our Synthetic Data for MT}
\label{sec:experiments}
We evaluate the quality of our synthetic data by analyzing model performance both before and after fine-tuning across multiple architectures (Sections \ref{sec:setup} and \ref{sec:results}), serving as a proxy for data quality. 
Furthermore, we compare our dataset to a web-crawled SOTA corpus (Section \ref{sec:hplt}), \textcolor{black}{investigate the effect of variable training data size (Section~\ref{sec:data_size})} and study the usefulness of MultiEuroparl (Section \ref{sec:finnish}).

\subsection{Experimental Setup}
\label{sec:setup}

\paragraph{Data} We focus on the translation direction from English into the low-resource language, 
as this is typically the more challenging scenario. For all experiments we use the synthetic data as training set, and the FLORES+ \cite{goyal2022flores} development and test sets for model selection and evaluation, respectively. 

\paragraph{Models} Since the languages under consideration are not linguistically similar, we train individual bilingual models for each target language and leave multilingual studies for future work. We experiment using the following models (more details are provided in Appendix~\ref{sec:train_regimes}):
\begin{itemize}[label={}, leftmargin=10pt]
    \item \textbf{Synthetic}: a transformer-base  model \citep{vaswani2017attention} trained on the synthetic data with MarianNMT \citep{junczys2018marian}.
    \item \textbf{OPUS-MT}: the best OPUS-MT model per language pair, based on the OPUS-MT Dashboard scores \citep{tiedemann2023opus}. The full list of the selected models is provided in Appendix~\ref{sec:opusmt_models}.  Each model is fine-tuned without modifying its original tokenizer. 
    \item \textbf{NLLB-200-distilled-1.3B}: the distilled 1.3B parameter NLLB-200 model \citep{nllbdistilled2022}. For fine-tuning NLLB, we used DeepSpeed \cite{deepspeed}. 
    \item \textbf{Llama-3.2-3B-Instruct}: the 3B parameter Llama-3.2 Instruct model \citep{grattafiori2024llama3herdmodels}. For the fine-tuned version, we adapt LoRA \citep{hu2022lora} using Unsloth \citep{unsloth}. 
\end{itemize}

All models are run on four 32 GB NVIDIA Volta V100 GPUs and take less than 9 hours to train.

\paragraph{Evaluation} We evaluate all models before and after fine-tuning. We report ChrF \citep{popovic2015chrf} as our main automatic metric, as it has been the standard metric for low-resource MT and it is shown to correlate more closely with human judgments than BLEU \citep{papineni-etal-2002}. We report COMET\footnote{We use the \texttt{\href{https://huggingface.co/Unbabel/wmt22-comet-da}{Unbabel/wmt22-comet-da}}  model.} \citep{rei2020comet} for all our experiments in Appendix \ref{app:comet}.

\textcolor{black}{We also evaluate GPT-4o on the full FLORES+ test set (in the pilot evaluation in Appendix \ref{sec:pilot_eval}, we used only 100 samples from the development set) and include it as a reference in our results.}

\subsection{Overall Results and Analysis} 
\label{sec:results}

Table~\ref{tab:chrf_results} summarizes the ChrF scores across three experimental conditions: off-the-shelf inference, fine-tuned training, and their performance differentials. We assessed the statistical significance of all the differences using paired Student’s \textit{t}-tests and paired bootstrap resampling (5000 iterations at 95\% confidence).

\paragraph{Effectiveness for Training from Scratch} The 60M parameter baseline, trained exclusively on our dataset, surpasses the out-of-the-box performance of billion-parameter models like NLLB and Llama for Basque, Scottish Gaelic, Icelandic and Georgian, while nearly matching them for Somali. This shows that our corpus is rich enough to train functional MT systems without any external pretraining or multilingual transfer. 
    
\paragraph{Impact on Fine-Tuning Pretrained Models} Fine-tuning consistently improves NLLB and Llama, confirming that the synthetic data is well-suited for adaptation. OPUS-MT also benefits from fine-tuning in five of seven cases; however, performance drops for Macedonian and Ukrainian, the two highest-resource low-resource pairs in our set. This suggests that when the model is trained on enough real parallel data, it ends up fitting too closely to the synthetic examples. 

\paragraph{Quality versus Usefulness}
Based on the results from the previous section, we observe high quality for Ukranian and Macedonian, medium quality for Basque and Icelandic, and noticeably lower quality for the rest. 
Yet, noisy does not mean useless. In fact, Table~\ref{tab:chrf_results} shows that the languages with the noisiest synthetic corpora also result in the largest downstream gains (Scottish Gaelic, Georgian and Somali). When the alternative is no data at all, quantity is better than quality. However, for mid-resource languages such as Macedonian and Ukrainian, cleaner text is already available beforehand \textcolor{black}{and multilingual pretrained models benefit from large quantities of data from closely related languages. Therefore,} additional synthetic data offers diminishing returns and mainly hurts the performance of these systems. The lower the resource level, the more tolerant MT training is to noise.

\paragraph{Challenges with General Purpose LLMs} Llama initially struggles (13-30 ChrF), reflecting its lack of inherent translation capability for low-resource languages. While adapter training yields substantial improvements (+11-29 ChrF), the model still underperforms compared to smaller, translation-specific models. This indicates that while synthetic data allows for adaptation, it cannot fully compensate for mismatches between the pretraining objective and the translation task itself.
In Llama's case, the 3B parameter scale appears unnecessarily large for this specific MT task, and leads to unnecessarily large fine-tuning times.

\paragraph{Competitiveness with GPT-4o}
\textcolor{black}{GPT-4o\footnote{Since GPT-4o’s training data is not public, FLORES+ may be included, making evaluation unfair.} is the best performing system for almost all language directions, however fine-tuned models like OPUS-MT-ft and NLLB-ft still offer competitive results, despite being much smaller. OPUS-MT-ft is far behind by 1-2 ChrF points in most languages and NLLB-ft even outperforms GPT-4o for Georgian. This indicates that while GPT-4o is a challenging system to beat, smaller and more efficient models can achieve comparable results with far fewer parameters.}
\textcolor{black}{These models close the performance gap using only our moderately-sized synthetic corpus ($\approx$2M sentences), a mere fraction of the vast data required to train a frontier model like GPT-4o. This underscores a key finding: a targeted strategy of generating high-quality data provides a powerful and practical pathway to SOTA performance. Our approach significantly lowers the computational and financial barriers, making high-quality MT for low-resource languages much more accessible.}

\paragraph{A practical recipe for exploiting fully–synthetic low-resource data}
These experiments point to a clear best practice:
\begin{enumerate}
    \itemsep0em 
    \item \textbf{Generate in bulk for truly low resource languages.} Prioritize volume over perfection, as even noisy data drives significant gains when no alternatives exist.
    \item \textbf{Fine-tune MT multilingual models.} NLLB-200 benefits consistently across all language directions. Our findings are consistent with previous research that finds that fine-tuning NLLB is among the best approaches for low-resource MT \cite{iyer-etal-2024-quality,zhu-etal-2024-multilingual,scalvinietal2025rethinking,tapo-etal-2025-bayelemabaga,de-gibert-etal-2025}.
    \item \textbf{Avoid general-purpose LLMs for low-resource MT.} Despite Llama's large gains, its inefficient computational costs and inferior translation performance confirm that translation-specific encoder-decoder models leverage synthetic data more effectively. 
\end{enumerate}



\subsection{Comparison with HPLT v2}
\label{sec:hplt}

To further assess our dataset's utility, we conduct comparative experiments against HPLT v2 \cite{de-gibert-etal-2024-new, burchell2025expandedmassivemultilingualdataset}, a 
"real" parallel corpus derived from web sources (Internet Archive\footnote{\href{https://archive.org/}{https://archive.org/}} and Common Crawl).\footnote{\href{https://commoncrawl.org/}{https://commoncrawl.org/}}

We train systems for three out of the four overlapping language pairs (English paired with Basque, Icelandic and Macedonian). This decision was motivated by the significant variation in HPLT v2 data sizes (see Appendix \ref{app:hplt_data}, Table \ref{tab:hplt_data}), with Ukrainian having approximately ten times more data than the others; therefore, we leave it out. We train three models: (1) the same synthetic baseline as described earlier (Synthetic), (2) a model trained on HPLT dataset (HPLT), and (3) a model trained on the concatenation of our synthetic dataset and the HPLT (HPLT + Synthetic). \textcolor{black}{All models follow the same architecture (transformer-base) and hyperparameters.}
All models are evaluated using ChrF on the same test set described previously. 
Table \ref{tab:hplt} reports the detailed ChrF scores.

\begin{table}[t]
    \centering
    \begin{tabular}{lrrrr}
    \toprule
    \multirow{2}{*}{Training Data} & \multicolumn{4}{c}{Language Pair} \\
     &  en-eu & en-is & en-mk \\
    \midrule
    Synthetic & 53.00 & 49.91 & 57.72 \\
    HPLT & 54.63 & 50.60 & 62.09 \\ \hdashline
$\Delta$ & +1.63* & +0.69* & +4.37*  \\        \midrule
    HPLT & 54.63 & 50.60 & 62.09 \\
    HPLT + Synthetic & \textbf{56.20}  &  \textbf{53.39} & \textbf{62.92}  \\ \hdashline
     $\Delta$ & +1.57* & +2.79* &  +0.83*    \\
    \bottomrule
    \end{tabular}
    \caption{ChrF scores for the comparison of our data with HPLT. }
    \label{tab:hplt}
\end{table}

\paragraph{Comparable Performance}
The models trained on our synthetic data alone perform on the same ballpark as the ones trained on the HPLT dataset, with an average difference of 2.23 ChrF points. The largest performance difference is observed for Macedonian, following a similar pattern as our experiments in the previous section.
These results demonstrate that our synthetic dataset is of sufficiently high quality to challenge real-world parallel corpora, even when trained from scratch.

\paragraph{Complementary when Combined}
Adding our corpus to HPLT yields the best overall performance across all language pairs, with significant improvements. This proves the effectiveness of our synthetic data in low-resource MT. The consistent increases suggest that our data introduces useful diversity and complements the HPLT dataset, as it represents previously unseen material.

\paragraph{Combined beats Transfer Learning}
If we compare these results with Table \ref{tab:chrf_results}, we can observe how training on the combined HPLT and synthetic datasets not only matches the performance of the fine-tuned NLLB for Basque, but surpasses it for both Icelandic and Macedonian, even though the Synthetic model is 21.6 times smaller. This highlights the power of data augmentation: enriching real-world corpora with high-quality synthetic data can outperform SOTA transfer learning approaches in low-resource settings.

\subsection{Effect of Training Data Size}
\label{sec:data_size}
\textcolor{black}{To assess the efficiency and scalability of our synthetic data approach, we train models with increasing fractions of the available synthetic data (10–100\%) by creating cumulative subsets. For each subset, we trained a Marian NMT model using the same tokenizer and identical hyperparameters to the baselines presented in Section \ref{sec:setup} to ensure comparability across runs, where data is the only changing variable.}

\begin{figure}
    \centering
    \includegraphics[width=1\linewidth]{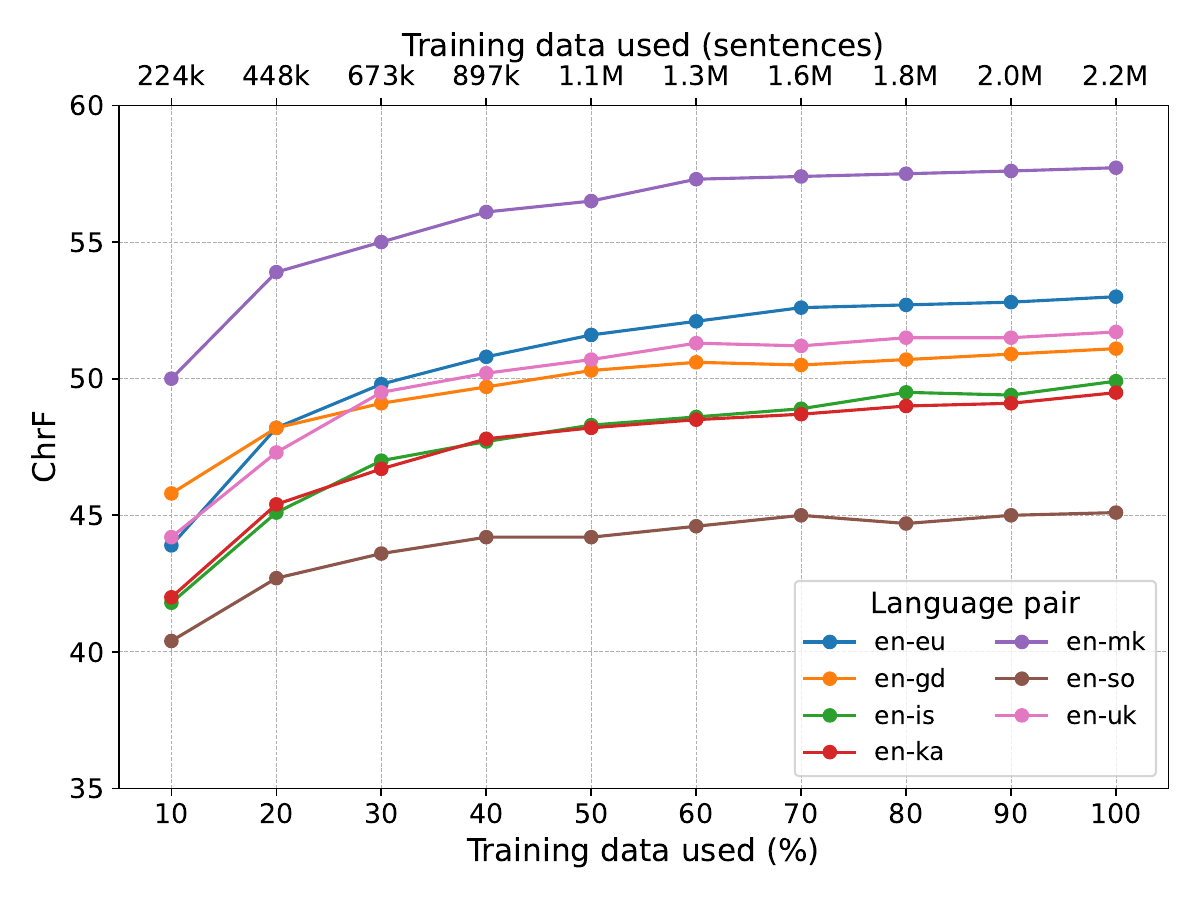}
    \caption{Learning curves with different data sizes.}
    \label{fig:curves}
\end{figure}

\paragraph{Effective Scaling with Synthetic Data}
\textcolor{black}{Figure \ref{fig:curves} shows that performance improves consistently as more data is used, confirming that additional synthetic data is beneficial across all language directions. However, the gains decrease after 50–60\% has been used. For example, English-Basque improves by +8.7 chrF from 10\% to 50\%, but only +1.4 additional points from 50\% to 100\%. 
This means that larger synthetic corpora brings gains but substantial improvements can already be achieved with a fraction of the full dataset.
This suggests that future research on synthetic data generation should prioritize data quality and diversity for greater benefits than further scaling alone.}

\subsection{Beyond English-Centric MT: The Finnish Use Case}
\label{sec:finnish}
To explore the multilingual potential of our expanded dataset via pivoting (Section \ref{sec:multieuroparl}) and move beyond English-centric translation, we train MT models for two additional language pairs: Finnish–Somali and Finnish-Ukrainian. \textcolor{black}{The data consists of 1 876 672 sentences for Finnish-Ukranian and 1 882 712 for Finnish-Somali.}

These languages were selected due to their prominence in Finland’s linguistic landscape, where Ukrainian and Somali are among the most widely spoken foreign languages, accounting for approximately 0.7\% and 0.5\% of the population, respectively \cite{OSF2024Population}. Developing high-quality models for these pairs is therefore both practical and relevant.

Our setup for this experiment is similar to the ones above.
We first evaluate the out-of-the-box capabilities of OPUS-MT and NLLB. Next, we train a synthetic baseline (transformer-base), and finally, we fine-tune OPUS-MT and NLLB. We exclude Llama fine-tuning from this stage, as previous results have shown that it consistently underperforms. Table \ref{tab:finnish} reports the results.

\begin{table}[t!]
    \centering
    \resizebox{\columnwidth}{!}{%
    \begin{tabular}{lrrrr}
    \toprule
    \multirow{2}{*}{Model} & \multicolumn{4}{c}{Language Pair} \\
     &  fi-so & so-fi & fi-uk & uk-fi \\
    \midrule
    Synthetic & 38.72 & 35.87 & 42.01 & 46.01 \\ \hline
    OPUS-MT  &26.31	&15.86	& \textbf{50.56} &	\textbf{55.24}\\
    OPUS-MT-ft  &41.09	& 37.50& 49.36 & 53.30 \\ \hdashline
    $\Delta$ &	 +14.78*	&+21.64*& -1.21* &	-1.94*	\\ \hline
    NLLB         & 40.33 & 39.66 & 45.59 & 47.78 \\
    NLLB-ft      & \textbf{42.21} & \textbf{42.31} & 47.61 & 51.62 \\ \hdashline
    $\Delta$     & +1.88* & +2.65* & +2.02* & +3.84* \\
    \bottomrule
    \end{tabular}
    }
    \caption{ChrF scores for Finnish-Somali and Finnish-Ukrainian translation.}
    \label{tab:finnish}

\end{table}



\paragraph{Usefulness of Pivoted Data} It is important to note that our synthetic baselines are weaker here than in previous experiments, providing greater headroom for fine-tuning improvements. 
OPUS-MT obtains clear gains from fine-tuning on synthetic data for the low-resource Finnish–Somali pair (+14.78 and +21.64 ChrF). However, for Finnish-Ukranian, fine-tuning does not improve. NLLB, which already exhibits a strong baseline, sees consistent gains across all directions. Overall, these results highlight the utility of synthetic data, particularly for low-resource language pairs.

\section{{SynOPUS}: a New Synthetic Parallel Corpus Repository}
The increasing adoption of LLMs in generating synthetic data underscores the growing need to systematically organize synthetic datasets. 
Although it is well known that many parallel datasets already contain MT content \cite{thompson-2024-shocking}, when synthetic data is intentionally produced, especially when involving significant financial or computational resources, proper  archiving are paramount for promoting reuse, ensuring transparency, and maximizing resource utility. Therefore, with the release of our dataset, we introduce {SynOPUS},\footnote{\href{https://opus.nlpl.eu/synthetic/}{https://opus.nlpl.eu/synthetic/}}
 a new repository for parallel synthetic datasets, i.e., data that has been (partially) generated by translating text into other languages using MT systems or LLMs. We invite the community to contribute with their own datasets.



\section{Conclusions}
In this work, we thoroughly studied the quality and usefulness of LLM-generated synthetic data for low-resource MT.
We presented a new synthetic corpus at document-level by forward translating Europarl, a parliamentary corpus, with GPT-4o. Then, we evaluated the resulting dataset both quantitatively and through human evaluation.
Furthermore, we investigated the usefulness of this dataset for low-resource MT by: (i) identifying the most effective strategy for training, (ii) comparing our dataset with the public HPLT dataset, (iii) extending our analysis beyond English-centric MT by generating a multi-way parallel corpus via pivoting through alignments to English, and \textcolor{black}{(iiii) studying the effect of varying training data size.}


Our study highlights a crucial and often overlooked opportunity: the ability to create valuable parallel resources for low-resource MT by leveraging widely available high-resource monolingual data. This challenges the traditional reliance on scarce real target-language data for data augmentation approaches, and opens new directions for scalable MT development.

For future work, we aim to explore optimal methods for combining real and synthetic data, as well as extending our experiments to the document-level and investigating the use of synthetic data for monolingual LLM pretraining.

\section*{Limitations}
\paragraph{Domain Bias} 
First and foremost, because of the origin of our source data, which is Europarl, a corpus compiled from parliamentary proceedings, the presented dataset belongs to a very specific domain. This implies that our models may suffer from domain bias and that any system trained on this data may not generalize well to informal, conversational, or domain-specific language; where linguistic style, vocabulary, and discourse structure differ significantly.

\paragraph{Language Coverage} While we focus on seven diverse languages (varied language families, linguistic typologies, written scripts), our approach relies on GPT-4o's ability to produce a certain language reasonably well. Even though we rely on the results of our pilot study (shown in Table \ref{tab:pilot_eval}), our method may not translate well to other languages.  Determining where the threshold lies, that is to say, how well a language must be supported for GPT-generated data to be viable; remains an open question.

\paragraph{Human Evaluation Scope} We aim to provide enough pointers to evaluate the quality of our dataset both numerically and qualitatively. However, our human evaluation is limited to a 100 samples per language pair due to a lack of resources.
Furthermore, we use Direct Assessment (DA), a widely accepted but increasingly outdated method. More recent evaluation approaches, such as Error Span Annotation (ESA)~\cite{kocmi2024error}, offer more fine-grained insights into translation errors, but were beyond our reach for this study.

\paragraph{Data Contamination of the Test Set}
Due to the the closed-source nature of GPT-4o, there is a risk of data contamination, since the model may have already seen our test set  (FLORES+) during pretraining. 
Recent studies \cite{mansurov2024data} have shown that distilled data may inherit intrinsic biases from the teacher model and this may have an impact on benchmark results.
\textcolor{black}{We note, however, that there is a strong domain mismatch between our source data (formal Europarl proceedings) and the FLORES+ benchmark (general encyclopedic text), which reduces the likelihood of simple memorization affecting results.}
While this risk is inherent to most widely used test sets and cannot be fully controlled, we acknowledge it here for transparency.

\paragraph{Data Contamination of the Source}
\textcolor{black}{GPT-4o pretraining data also likely includes the Europarl corpus. 
This means our experiments could be affected by data contamination, in the sense that the model may have had indirect prior exposure to the underlying content and domain, even if not in our low-resource target languages. 
Because our setup requires cross-lingual generation into languages that are not present in Europarl, direct memorization is unlikely. 
Still, there remains a risk of overestimating translation quality, which should be kept in mind when interpreting our results.}

\section*{Ethical Considerations}

\paragraph{Hallucinated Content} LLMs are known to generate hallucinated content, outputs that are fluent and well-formed but factually incorrect or irrelevant~\cite{vazquez2025semeval}.
This phenomenon is a risk in iself,  as it can introduce noise and propagate misinformation in downstream MT models. In our case, we observed that in some cases, the model disregards the input and instead generates a response similar to, ``You have been trained on data up until October 2023'' in the target language. This issue is most prevalent in Georgian, with around 9,000 cases, and Ukrainian, with approximately 4,000 cases. For these languages, we removed each line containing the string ``2023''. While such hallucinations appear to be an intrinsic limitation of current LLMs, they highlight the need for careful post-processing and validation when using synthetic data.

\paragraph{Reproducibility}  Since our synthetic data is generated using a closed-source LLM, the exact reproduction of our work is not possible. To mitigate this, we publicly release the generated dataset along with all preprocessing scripts and training code.

\textcolor{black}{\paragraph{Cost-Benefit Trade-off}  Our empirical results demonstrate that augmenting training data with high-quality LLM-generated translations improve translation performance for low-resource languages, outperforming existing baselines. This benefit is valuable in contexts where existing parallel training data is scarce or even unavailable. However, the cost of generating such translations with LLMs is significant both in terms of compute and financial expense, and may therefore be unreasonable for many groups. For example, generating the synthetic data used in our work costed approximately \$5\,000. While the improvements in translation quality may justify the use of LLMs for data generation, future work should explore more cost-efficient methods for synthetic data generation. Promising directions could include e.g., distilling larger models into smaller ones and selective data augmentation to reduce the volume of unnecessary synthetic data while preserving improvements in performance.}

\vspace{5mm}
ChatGPT was used to assist code development for this project.

\section*{Acknowledgments}

\textcolor{black}{First and foremost, we would like to sincerely thank our annotators for their valuable contributions.
Ander González Docasal, Harritxu Gete Ugarte, and Nerea Mandiola Solozabal for the Basque annotation;
Helga Hilmisdóttir and Reynir Eggertsson for Icelandic;
Artur Voit-Antal and Yana Matyash for Ukrainian;
Biljana Stojanovska for Macedonian;
and Elene Kavteladze for Georgian.}

We acknowledge OpenAI for their generous provision of \$5,000 in API credits, which were instrumental in generating the translations that comprise the datasets used in this work.

\textcolor{black}{This project has received funding from the European Union’s Horizon Europe programme (GA No 101070350) and from UK Research and Innovation (UKRI) under the UK government’s Horizon Europe funding guarantee (GA No 10052546).
This work was also supported by the Digital Europe Programme under grant agreement No 101195233.
and by the GreenNLP project, which is funded by the Research Council of Finland.
Tiancheng Hu is supported by Gates Cambridge Trust (grant OPP1144 from the Bill \& Melinda Gates Foundation)}.


\bibliography{anthology, custom}

\appendix
\newpage
\onecolumn
\section{Pilot Evaluation Results}
\textcolor{black}{Table \ref{tab:pilot_eval} reports the ChrF scores of GPT-4o and EMMA on low-resource MT for a 100 random sample of the FLORES+ development set. We compare performance at the sentence (Sent.) and chunk (Chunk) level. For EMMA, both zero-shot (Sent./0) and three-shot (Sent./3) settings are reported. The best OPUS-MT model is included as a reference. There is no OPUS-MT model for Aragonese and Aranese.}

\textcolor{black}{We do not evaluate our synthetic baseline models on these sample sentences, as the development set was used during training. Evaluating on this data risks overfitting and leads to optimistically biased performance estimates, which may not reflect true generalization.}

\label{sec:pilot_eval}
\begin{table}[h!]
    \centering
    \begin{adjustbox}{width=0.7\linewidth}
    \normalsize
    \input{tabs_and_figs/pilot_eval}
    \end{adjustbox}
    \caption{ChrF scores of the evaluation of GPT-4o and EMMA on low-resource MT. Highlighted rows correspond to the final set of selected languages for our study.}
    \label{tab:pilot_eval}
\end{table}

\newpage

\newpage
\onecolumn 
\section{Human Evaluation Annotation Guidelines}
\textcolor{black}{In Figure \ref{fig:guidelines_instructions}, we provide an exact copy of the annotation guidelines given to the annotators.}
\label{sec:guidelines}
\begin{figure*}[!th]
{

    \textbf{Introduction}\\
    You will be asked to evaluate the quality of machine-translated (MT) sentences by comparing each one directly to its human-written original sentence (the source sentence). You will assign a score, based on how well the translation preserves meaning, fluency, and naturalness. This is what is known as Direct Assessment (DA, Graham et al., 2013). DA elicits human assessments of translation adequacy on an analogue rating scale (0–100), where human assessors are asked to rate how adequately the APE system output expresses the meaning of the human reference translation (Bojar et al., 2017).
    In this annotation project, you will be shown 100 samples of source-hypothesis pairs. Your task is to evaluate each translation pair through DA.\\

    \textbf{Annotation Guidelines}
    \vspace{-0.5em}
    
    \begin{enumerate}
    \setlength\itemsep{-0.3em}
    \item Carefully read the sentence pair. Try to understand the intended meaning of the source.
    \item Evaluate whether the sentences are parallel or not.  Compare the MT sentence with the source. Does the MT output preserve the key meaning of the source sentence?
    \item Evaluate whether the target sentence contains fluency mistakes.  Is the MT sentence grammatically correct? Are there any strange phrases, broken structure, or missing words?
    \item Decide the score based on the scoring scale below.
    \item Ensure that you double-check your annotations prior to moving to the next example.  Re-read both source and translation. Does the score reflect meaning and fluency? Were you consistent with your previous scores? Adjust the score if needed to maintain fairness and consistency.
    \end{enumerate}

    \textbf{Scoring scale}\\
    \vspace{-0.5em}
    Use the full range of the scale. Do not be afraid to give very low or very high scores when appropriate.\\

    \centering
    \begin{tabular}{rl}
\textbf{Score} & \textbf{Interpretation} \\
100 & Perfect: grammatically flawless, fluent, and semantically identical to the source.\\
85–99 & Excellent: small stylistic or fluency issues; all meaning preserved.\\
70–84 & Good: mostly fluent; minor issues in grammar, wording, or slight meaning distortion.\\
50–69 & Acceptable: understandable, but multiple issues with grammar, style, or partial meaning loss.\\
30–49 & Poor: hard to understand, major meaning lost, broken grammar.\\
1–29 & Very poor: barely comprehensible or mostly wrong meaning.\\
0 & Incomprehensible: completely unrelated, meaningless, or unreadable.\\
    \end{tabular}
}

    \caption{Annotation guidelines: Instructions.}
    \label{fig:guidelines_instructions}
\end{figure*}
\twocolumn
\newpage
\section{Training Regimes}
\label{sec:train_regimes}
\begin{itemize}
    \item \textbf{Synthetic}: We employ a shared 32k SentencePiece \cite{kudorichardson-2018-sentencepiece} vocabulary trained on the synthetic corpus; other settings follow the original Transformer-base recipe. 
    Mini-batch fitting is enabled to optimize memory usage. Validation every 2500 updates checks perplexity. Early-stopping is employed on the development set, with a patience of 10. 
    \item \textbf{OPUS-MT-ft}: We fine-tune each model without modifying its original tokenizer; appropriate language tags are prefixed at train and test time for multilingual models. Mini-batch fitting is enabled to optimize memory usage. Validation every 500 updates checks perplexity. Early-stopping is employed on the development set, with a patience of 20. 
    \item \textbf{NLLB-200-distilled-1.3B-ft}: We fine-tune the NLLB-distilled-1.3B model with DeepSpeed on four V100 GPUs in FP16 mixed precision. Training uses a per-GPU batch size of 32 sentences, a maximum sequence length of 128 tokens, the Adam optimiser with a 1 $\times 10^{-4}$ learning rate, and runs for up to four epochs. DeepSpeed ZeRO-1 is used for basic tensor sharding; everything else is left on-GPU. Early-stopping is employed on the development set, with a patience of 5. 
    \item \textbf{Llama-3.2-3B-Instruct-ft}: We adapted the model with LoRA using the Unsloth framework. We used the quantized 4-bit version of the model, applying LoRA adapters, and we used with prompts designed to mimic a professional translator's task using Unsloth’s template system. Training was done using SFTTrainer with fp16 mixed precision, gradient accumulation, and 50k training steps with effective batch size of 16 utterances. 

\end{itemize}
\newpage
\section{OPUS-MT Models selected for fine-tuning}
\label{sec:opusmt_models}
We select the best available OPUS-MT model based on the OPUS-MT Dashboard \cite{tiedemann2023opus}, by looking at the BLEU score on the FLORES+ dataset.
\begin{itemize}
\item en-eu: \href{https://huggingface.co/HPLT/translate-en-eu-v1.0-hplt_opus}{translate-en-eu-v1.0-hplt\_opus}
\item en-gd: \href{https://object.pouta.csc.fi/Tatoeba-MT-models/deu+eng+fra+por+spa-ine/opusTCv20230926max50+bt+jhubc_transformer-big_2024-05-30.zip}{deu+eng+fra+por+spa-ine/tf-big}
\item en-is: \href{https://huggingface.co/HPLT/translate-en-is-v1.0-hplt_opus}{translate-en-is-v1.0-hplt\_opus}
\item en-mk: \href{https://object.pouta.csc.fi/Tatoeba-MT-models/deu+eng+fra+por+spa-sla/opusTCv20230926max50+bt+jhubc_transformer-big_2024-05-30.zip}{deu+eng+fra+por+spa-sla/tf-big}
\item en-so: \href{https://object.pouta.csc.fi/Tatoeba-MT-models/deu+eng+fra+por+spa-afa/opusTCv20230926max50+bt+jhubc_transformer-big_2024-05-29.zip}{deu+eng+fra+por+spa-afa/tf-big}
\item en-uk: \href{https://object.pouta.csc.fi/Tatoeba-MT-models/eng-zle/opusTCv20210807+bt_transformer-big_2022-03-13.zip}{eng-zle/tf-big}
\item en-ka: \href{https://object.pouta.csc.fi/Tatoeba-MT-models/deu+eng+fra+por+spa-cau/opusTCv20230926max50+bt+jhubc_transformer-big_2024-05-30.zip}{deu+eng+fra+por+spa-cau/tf-big}
\item fi-so: \href{https://object.pouta.csc.fi/Tatoeba-MT-models/mul-mul/opusTCv20230926+bt+jhubc_transformer-big_2024-08-17.zip}{mul-mul/tf-big}
\item fi-uk: \href{https://object.pouta.csc.fi/Tatoeba-MT-models/fin-zle/opusTCv20210807+xb+bt+pft+pbt_transformer-big_2022-04-27.zip}{fin-zle/tf-big}
\item so-fi: \href{https://object.pouta.csc.fi/Tatoeba-MT-models/afa-fiu/opus-2021-02-12.zip}{afa-fiu/tf-base}
\item uk-fi: \href{https://object.pouta.csc.fi/Tatoeba-MT-models/zle-fin/opusTCv20210807+xb+bt+pft+pbt_transformer-big_2022-04-18.zip}{zle-fin/tf-big}

\end{itemize}
\newpage
\onecolumn

\section{COMET scores for MT training}
\label{app:comet}

\textcolor{black}{We report the COMET scores for all our experiments to provide a more comprehensive evaluation.
Table \ref{tab:comet_results} shows the COMET scores equivalent to Table \ref{tab:chrf_results} for our fine-tuning experiments. Table \ref{tab:hplt_comet} shows the COMET scores equivalent to Table \ref{tab:hplt} for our comparison with HPLT. Finally, Table \ref{tab:finnish_comet} shows the COMET scores equivalent to Table \ref{tab:finnish} for our study on Finnish-centric translation.}

\begin{table*}[ht]
    \centering
    \begin{tabular}{lrrrrrrrr}
    \toprule
      \multirow{2}{*}{Model} & \multicolumn{7}{c}{Language Pair} &  \multirow{2}{*}{\parbox[t]{1.5cm}{$\#$ Params}} \\
   & en-eu & en-gd & en-is & en-ka & en-mk & en-so & en-uk &  \\
    \midrule
    Synthetic & 81.51 & 78.04 & 80.16 & 80.72 & 82.24 & 78.15 & 78.89 & 60.6M \\ \hline
   OPUS-MT & 83.27 & 71.30 & 79.69 & 69.09 & 87.34 & 77.06 & \textbf{89.02} & \multirow{2}{*}{191.6M}  \\
    OPUS-MT-ft & 84.15 & 79.30 & 83.21 & 81.60 & 86.19 & 79.34 & 87.61 & \\ \hdashline
    $\Delta$ & +0.88 & +8.00 & +3.52 & +12.51 & -1.15 & +2.28 & -1.41 \\ \hline
   NLLB & 84.55 & 78.73 & 82.06& 80.49&87.45& 80.06 & 87.21& \multirow{2}{*}{1.3B}   \\
    NLLB-ft & \textbf{86.84} & \textbf{79.43} & \textbf{85.36} & \textbf{86.94} & \textbf{88.74} & \textbf{80.90}&88.14&    \\ \hdashline
    $\Delta$  & +2.29 & +0.7 & +3.3 & +6.45 & +1.29 & +0.84 & +0.93  \\\hline   
    Llama & 40.94 & 44.61 & 36.96 & 33.68 & 42.50 & 43.79 & 50.86 & \multirow{2}{*}{3B} \\
    Llama-ft & 70.00 & 75.92 & 76.63 & 54.53 & 82.09 & 76.67 & 80.80 &     \\ \hdashline
    $\Delta$ & +29.06 & +31.31 & +39.67 & +20.85 & +39.59 & +32.88 & +29.94 \\ 
    \midrule
    GPT-4o &86.65 & 80.11 & 86.70 & 85.76 & 90.04 & 80.67 & 91.13 & \\
    \bottomrule
    \end{tabular}
    \caption{COMET scores for our fine-tuning experiments.}
    \label{tab:comet_results}
\end{table*}

\begin{table}[h]
    \centering
    \begin{tabular}{lrrrr}
    \toprule
    \multirow{2}{*}{Training Data} & \multicolumn{4}{c}{Language Pair} \\
     &  en-eu & en-is & en-mk \\
    \midrule
    Synthetic & 81.51 & 80.16 & 82.24 \\
    HPLT & 82.47 & 78.09 & 85.38 \\ \hdashline
$\Delta$ & +0.96 & -2.07 &  + 3.14\\        \midrule
    HPLT & 82.47  & 78.09 & 85.38 \\
    HPLT + Synthetic & \textbf{84.53}  &  \textbf{82.82} & \textbf{86.96}  \\ \hdashline
     $\Delta$ & +2.06 & +4.73 & +1.58    \\
    \bottomrule
    \end{tabular}
    \caption{COMET scores for the comparison of our data with HPLT.}
    \label{tab:hplt_comet}
\end{table}

\begin{table}[bh!]
    \centering
    \begin{tabular}{lrrrr}
    \toprule
    \multirow{2}{*}{Model} & \multicolumn{4}{c}{Language Pair} \\
     &  fi-so & so-fi & fi-uk & uk-fi \\
    \midrule
    Synthetic & 75.09 & 66.55 & 76.89 & 77.36 \\ \hline
    OPUS-MT  &54.06	&34.33	& \textbf{89.07} &	\textbf{88.01}\\
    OPUS-MT-ft  &76.72	& 68.32& 87.77 & 87.10 \\ \hdashline
    $\Delta$ &	+22.66 & +33.99 & -1.30 & -0.91		\\ \hline
    NLLB         & 77.35 & 76.37 & 85.17 & 84.40 \\
    NLLB-ft      & \textbf{78.60} & \textbf{78.62} & 87.09 & 87.04 \\ \hdashline
    $\Delta$     & +1.25 & +2.25 & +1.92 & +2.64  \\
    \bottomrule
    \end{tabular}
    \caption{COMET scores for Finnish-Somali and Finnish-Ukrainian translation.}
    \label{tab:finnish_comet}

\end{table}

\newpage
\section{HPLT Data Sizes}
\label{app:hplt_data}
\textcolor{black}{We report the total amount of sentences of the HPLT v2 dataset in Table \ref{tab:hplt_data} for the overlapping language pairs with our selected languages.}
\begin{table}[h]
    \centering
    \begin{tabular}{lcccc}
    \toprule
     &  en-eu & en-is & en-mk & en-uk \\
    \midrule
    n. sentences & 1 491 873&2 694 541 &3 991 617 & 25 125 019 \\
    \bottomrule
    \end{tabular}
    \caption{Data sizes in amount of sentences in the HPLT v2 dataset.}
    \label{tab:hplt_data}
\end{table}

\end{document}

%% file: tabs_and_figs/pilot_eval.tex
\begin{tabular}{ll|r|rrrr}
\toprule
& & OPUS-MT & \multicolumn{2}{c}{GPT-4o} & \multicolumn{2}{c}{EMMA} \\
Language & Code & Sent. & Sent. & Chunk & Sent./3 & Sent./0 \\
\midrule
Aragonese & arg\_Latn & - & 43.9 & 47.5 & 38.1 & 39.8 \\
Aranese & arn\_Latn & - & 41.2 & 45.4 & 16.7 & 27.6 \\
Armenian & hye\_Armn & 47.5 & 56.3 & 56.5 & 45.7 & 48.7 \\
Asturian & ast\_Latn & 59.6 & 59.6 & 62.8 & 52.7 & 53.3 \\
Bashkir & bak\_Cyrl & 40.6 & 51.4 & 52.2 & 24.2 & 29.1 \\
\rowcolor[gray]{0.85}Basque & eus\_Latn & 55.0 & 57.3 & 60.4 & 41.5 & 44.0 \\
Belarusian & bel\_Cyrl & 44.4 & 46.6 & 49.2 & 39.0 & 37.9 \\
Bosnian & bos\_Latn & 58.8 & 63.9 & 65.4 & 51.6 & 53.3 \\
Catalan & cat\_Latn & 67.4 & 67.9 & 69.8 & 56.7 & 56.5 \\
Crimean Tatar & crh\_Latn & 35.9 & 37.6 & 40.0 & 13.7 & 23.2 \\
Croatian & hrv\_Latn & 61.5 & 60.9 & 62.2 & 50.9 & 51.2 \\
Esperanto & epo\_Latn & 59.7 & 63.2 & 65.6 & 60.8 & 60.5 \\
Friulian & fur\_Latn & 49.9 & 45.7 & 50.0 & 45.4 & 43.8 \\
Galician & glg\_Latn & 62.5 & 63.3 & 66.2 & 55.7 & 55.9 \\
\rowcolor[gray]{0.85}Georgian & kat\_Geor & 42.7 & 51.3 & 42.4 & 45.3 & 45.9 \\
Hebrew & heb\_Hebr & 61.3 & 58.1 & 59.2 & 41.1 & 47.0 \\
\rowcolor[gray]{0.85}Icelandic & isl\_Latn & 53.0 & 55.6 & 59.3 & 41.1 & 41.6 \\
Irish & gle\_Latn & 60.5 & 61.3 & 61.6 & 48.9 & 49.3 \\
Ligurian & lij\_Latn & 43.9 & 36.2 & 40.1 & 34.5 & 35.3 \\
Limburgish & lim\_Latn & 36.5 & 43.7 & 44.6 & 28.3 & 29.8 \\
Lombard & lmo\_Latn & 34.1 & 35.3 & 38.8 & 25.3 & 28.4 \\
Luxembourgish & ltz\_Latn & 55.5 & 59.4 & 60.9 & 51.5 & 49.9 \\
\rowcolor[gray]{0.85}Macedonian & mkd\_Cyrl & 64.5 & 65.3 & 64.2 & 55.2 & 56.8 \\
Northern Uzbek & uzn\_Latn & 12.7 & 59.3 & 60.7 & 30.5 & 44.4 \\
Occitan & oci\_Latn & 64.0 & 67.5 & 67.0 & 51.7 & 46.0 \\
Sardinian & srd\_Latn & 50.1 & 43.4 & 46.2 & 40.0 & 48.8 \\
\rowcolor[gray]{0.85}Scottish Gaelic & gla\_Latn & 42.6 & 52.4 & 55.7 & 44.5 & 45.9 \\
Serbian & srp\_Cyrl & 63.0 & 64.1 & 65.0 & 35.5 & 46.7 \\
Sicilian & scn\_Latn & 39.9 & 45.0 & 48.7 & 43.4 & 43.8 \\
\rowcolor[gray]{0.85}Somali & som\_Latn & 44.2 & 47.6 & 53.0 & 42.7 & 38.6 \\
Tatar & tat\_Cyrl & 43.0 & 53.6 & 54.9 & 21.1 & 33.9 \\
Tosk Albanian & als\_Latn & 53.9 & 61.3 & 63.1 & 44.7 & 54.4 \\
Turkish & tur\_Latn & 62.8 & 66.1 & 67.1 & 36.8 & 34.5 \\
Turkmen & tuk\_Latn & 42.6 & 55.1 & 54.8 & 22.3 & 25.8 \\
\rowcolor[gray]{0.85}Ukrainian & ukr\_Cyrl & 60.1 & 60.1 & 62.1 & 48.7 & 49.4 \\
Uyghur & uig\_Arab & 37.1 & 38.1 & 36.8 & 30.2 & 31.1 \\
Venetian & vec\_Latn & 44.6 & 49.5 & 52.0 & 34.7 & 37.1 \\
Welsh & cym\_Latn & 64.9 & 73.3 & 73.6 & 59.4 & 59.8 \\
Yiddish & ydd\_Hebr & 0.0 & 40.5 & 42.6 & 41.4 & 51.5 \\
\midrule
Average &  & 49.2 & 53.9 & 55.6 & 40.8 & 43.6 \\
\bottomrule
\end{tabular}